# What Counterfactuals Can Be Tested

**Ilya Shpitser, Judea Pearl**
Cognitive Systems Laboratory
Department of Computer Science
University of California, Los Angeles
Los Angeles, CA. 90095
{ilyas, judea}@cs.ucla.edu

## Abstract

Counterfactual statements, e.g., "my headache would be gone had I taken an aspirin" are central to scientific discourse, and are formally interpreted as statements derived from "alternative worlds". However, since they invoke hypothetical states of affairs, often incompatible with what is actually known or observed, testing counterfactuals is fraught with conceptual and practical difficulties. In this paper, we provide a complete characterization of "testable counterfactuals," namely, counterfactual statements whose probabilities can be inferred from physical experiments. We provide complete procedures for discerning whether a given counterfactual is testable and, if so, expressing its probability in terms of experimental data.

## 1 Introduction

Human beings organize their knowledge of the world in terms of causes-effect relationships, because many of the practical questions they face are causal in nature. Counterfactuals are an example of causal questions which abound both in everyday discourse, as well as in empirical science, medicine, law, public policy, economics, and so on.

A counterfactual is simply a 'what if' question – it involves evidence about an existing state of the world, e.g. "I have a headache", and a question about an alternative, hypothetical world, where the past is modified in some way, e.g., "what if I had taken aspirin?". To formalize such questions, we need a framework that can seamlessly integrate the notions of evidence and 'world alteration,' such as that provided by structural causal models [Pearl, 2000a]. Such models are represented by a graph called a *causal diagram*, where the vertices $\mathbf{V}$ are variables of interest, directed edges represent functional relationships, and bidirected edges are spurious dependencies emanating from variables not included in the analysis, over which a probability distribution $P(\mathbf{U})$ is assumed to be defined. This distribution, together with the functional relationships among the variables defines a unique joint probability distribution $P(\mathbf{V})$ over observable variables $\mathbf{V}$, which governs statistical data obtained in observational studies.

The results of *observing* some aspect $e$ of the current state of affairs leads to conditional distributions $P(\mathbf{V}|e)$ and $P(\mathbf{U}|e)$. In contrast, the result of hypothetically establishing $\mathbf{x}$ is represented by an *interventional* distribution $P(\mathbf{V}|do(\mathbf{x}))$ or $P_{\mathbf{x}}(\mathbf{V})$, where $do(\mathbf{x})$ stands for hypothetically forcing variables $\mathbf{X}$ to attain values $\mathbf{x}$ regardless of the factors that influence $\mathbf{X}$ in the model while leaving all other functional relationships unaltered. A variable $Y$ affected by an intervention $do(\mathbf{x})$ is changed into a *counterfactual variable* and is denoted by $Y_{\mathbf{x}}$.[1]

To represent a 'what if $\mathbf{X}$ were $\mathbf{x}$' question, we assume the state of knowledge $P(\mathbf{U}|e)$ induced by the observations $e$, and ask for the consequences of taking the atomic action $do(\mathbf{x})$, where actions and observations can potentially be in conflict. In our framework, this corresponds to expressions of the form $P(Y_{\mathbf{x}}|e)$. This way of mathematizing counterfactuals was first proposed in [Balke & Pearl, 1994b], [Balke & Pearl, 1994a]. In addition, [Balke & Pearl, 1994b] proposed a method for evaluating expressions like the above when all parameters of a causal model are known. In practice, however, complete knowledge of the model is too much to ask for; the functional relationships as well as the distribution $P(\mathbf{U})$ are not known exactly, though some of their aspects can be inferred from the observable distribution $P(\mathbf{V})$.

Evaluating causal queries given this partial state of knowledge is a subtle problem known as *identification* [Pearl, 2000a]. A well studied version of this problem is comput-ing *causal effects*, or expressions of the form $P_\mathbf{x}(Y)$, given $P$ and the causal diagram $G$. This version of the identification problem has received considerable attention in the last 15 years, with partial results found in [Spirtes, Glymour, & Scheines, 1993], [Pearl & Robins, 1995], [Pearl, 1995], [Kuroki & Miyakawa, 1999], [Tian & Pearl, 2002], and was finally closed in general graphical models in [Huang & Valtorta, 2006], [Shpitser & Pearl, 2006b], [Shpitser & Pearl, 2006a].

---

[1] In practice, attempts to physically intervene on one variable may have unintended side effects. Still, a semantics based on ideal, atomic interventions provides a useful abstraction (similar to "derivative" in calculus), with the help of which the impact of compound interventions, side effects included, can be analyzed with mathematical precision.

The problem with counterfactual queries like $P(Y_\mathbf{x}|e)$ is even more severe. Since actions and evidence can stand in logical contradiction, no experimental setup exists which would emulate both the evidence $e$ and the action $x$. For example, no experimental setup can reveal to us the percentage of deaths that could be avoided among people who received a given treatment, had they not taken the treatment. We simply cannot perform an experiment where the same person is both given and not given treatment. Mathematically, this means that it is unclear whether counterfactual expressions like $P(Y_\mathbf{x}|e)$, with $e$ and $\mathbf{x}$ incompatible, can be estimated consistently even if we are given the results of all possible experiments (represented by the set $P_* = \{P_\mathbf{x}|$ where $\mathbf{x}$ is a value assignment of $\mathbf{X} \subseteq \mathbf{V}\}$ [Pearl, 2000a]).

Some basic results on evaluating counterfactuals are known. For instance, a simple algebraic trick shows that $P(Y_x|x')$ is experimentally identifiable (i.e., computable from $P_*$) if $X$ is a binary variable, regardless of the underlying graph. On the other hand, the counterfactual represented by $P(Y_x, Y_{x'})$, named 'probability of necessity and sufficiency' in [Pearl, 2000a], is known to not be experimentally identifiable [Avin, Shpitser, & Pearl, 2005], unless additional assumptions can be brought to bear (e.g., monotonicity [Pearl, 2000a]). In this paper we explore testability of counterfactuals relative to scientific knowledge expressed in the form of missing links in the underlying graph. The sensitivity of the tested quantities to this extra knowledge can be assessed using the bounding method of [Balke & Pearl, 1994a].

A complete proof system for reasoning about causal and counterfactual quantities was given in [Halpern, 2000]. While such a system is, in principle, powerful enough to evaluate any identifiable counterfactual expression, it lacks a proof guiding method which guarantees termination in a reasonable amount of time. Furthermore, such a system would not provide a graphical characterization of identification, and much of human knowledge, as we postulate, is stored in graphical form. To the best of the authors' knowledge, no general algorithms for counterfactual identification exist in the literature.

In this paper, we present a structure called the *counterfactual graph*, which stands in the same relation to a counterfactual query that the causal graph does to a causal query. In other words, this graph displays independencies between counterfactual variables, in those hypothetical worlds that are invoked by the query. We use the counterfactual graph to give a complete graphical characterization of those counterfactuals which can be identified from experiments, and provide complete algorithms which can express all identifiable counterfactuals in terms of experimental data.

## 2 Notation and Definitions

In this section we review the mathematical machinery of causal reasoning, and introduce counterfactual distributions as well-defined objects obtained from causal models.

A probabilistic causal model is a tuple $M = \langle \mathbf{U}, \mathbf{V}, \mathbf{F}, P(\mathbf{U}) \rangle$, where $\mathbf{V}$ is a set of observable variables, $\mathbf{U}$ is a set of unobservable variables distributed according to $P(\mathbf{U})$, and $\mathbf{F}$ is a set of functions. Each variable $V \in \mathbf{V}$ has a corresponding function $f_V \in \mathbf{F}$ that determines the value of $V$ in terms of other variables in $\mathbf{V}$ and $\mathbf{U}$. The distribution on $\mathbf{V}$ induced by $P(\mathbf{U})$ and $\mathbf{F}$ will be denoted $P(\mathbf{V})$.

The induced graph $G$ of a causal model $M$ contains a node for every element in $\mathbf{V}$, a directed edge between nodes $X$ and $Y$ if $f_Y$ possibly uses the values of $X$ directly to determine the value of $Y$, and a bidirected edge between nodes $X$ and $Y$ if $f_X$ and $f_Y$ both possibly use the value of some variable in $\mathbf{U}$ to determine their values. In this paper we consider *recursive* causal models, those models which induce acyclic graphs. We will use abbreviations $Pa(.)_G, Ch(.)_G, An(.)_G, De(.)_G$ to denote the set of parents, children, ancestors and descendants of a given node in $G$.

An action $do(\mathbf{x})$ modifies the functions associated with $\mathbf{X}$ from their normal behavior to outputting constant values $\mathbf{x}$. The result of an action $do(\mathbf{x})$ on a model $M$ is a *submodel* which we denote by $M_\mathbf{x}$. Because the nodes $\mathbf{X}$ are now constant, the graph induced by $M_\mathbf{x}$ is $G \setminus \mathbf{X}$. We denote the event "variable $Y$ attains value $y$ in $M_\mathbf{x}$" by the shorthand "$y_\mathbf{x}$".

Consider a conjunction of events $\gamma$ equal to $y^1_{\mathbf{x}^1} \wedge ... \wedge y^k_{\mathbf{x}^k}$ in some model $M$. If all subscripts $\mathbf{x}^i$ are the same and equal to $\mathbf{x}$, this $\gamma$ merely corresponds to value assignments to a set of variables in a submodel $M_\mathbf{x}$. The probability of this assignment is then $P(\gamma) = P_\mathbf{x}(y^1, ..., y^k)$ which can be easily computed from $P_\mathbf{x}$. But what if the subscripts are not the same, and possibly force conflicting values to the same variable? A natural way to interpret our conjunction in this case is to consider all submodels $M_{\mathbf{x}^1}, ..., M_{\mathbf{x}^k}$ at once, and compute the joint probability over the counterfactual variables in those submodels induced by $\mathbf{U}$, the set of exogenous variables all these submodels have in common. The probability of our conjunction is then given by $P(\gamma) = \sum_{\{\mathbf{u}|\mathbf{u} \models \gamma\}} P(\mathbf{u})$ where $\mathbf{u} \models \gamma$ is taken to mean that each variable assignment in $\gamma$ holds true in the corresponding submodel of $M$ when the exogenous variables $\mathbf{U}$ assume values $\mathbf{u}$. In this way, $P(\mathbf{U})$ induces a distribution on all counterfactual variables in $M$. In this paper, we will represent counterfactual utterances by joint distributions such as $P(\gamma)$ or conditional distributions such as $P(\gamma|\delta)$, where $\gamma$ and $\delta$ are conjunctions of counterfactual events. See [Pearl, 2000a] for an extensive discussion of counterfactuals, and their probabilistic representation used in this paper.

We are interested in finding out when queries like $P(\gamma)$ can be computed from $P_*$, the set of all interventional distributions, and when they cannot. To get a handle on this question, we turn to the notion of identifiability, which has been successfully applied to similar questions involving causal effects $P_\mathbf{x}(\mathbf{Y})$ [Pearl, 2000a].

**Definition 1 (identifiability)** *Consider a class of models $\mathbf{M}$ with a description $T$, and objects $\phi$ and $\theta$ computable from each model. We say that $\phi$ is $\theta$-identified in $T$ if $\phi$ is uniquely computable from $\theta$ in any $M \in \mathbf{M}$.*

If $\phi$ is $\theta$-identifiable in $T$, we write $T, \theta \vdash_{id} \phi$. Otherwise, we write $T, \theta \not\vdash_{id} \phi$. The above definition leads naturally to a way to prove non-identifiability.

**Lemma 1** *Let $T$ be a description of a class of models $\mathbf{M}$. Assume there exist $M^1, M^2 \in \mathbf{M}$ that share objects $\theta$, while $\phi$ in $M^1$ is different from $\phi$ in $M^2$. Then $T, \theta \not\vdash_{id} \phi$.*

In the remainder of the paper, we will construct an algorithm which, for any $T = G$, will identify $\phi = P(\gamma|\delta)$ (with $\delta$ possibly empty) from $\theta = P_*$, and prove that whenever the algorithm fails, the original query is not identifiable using Lemma 1.

## 3 The Counterfactual Graph

Solutions to the causal effect identification problem rely on judging independencies among random variables in the same submodel $M_\mathbf{x}$ using d-separation [Pearl, 1988] in the causal graph $G \setminus \mathbf{X}$. If we are dealing with a counterfactual $\gamma$, more than one submodel is mentioned. Nevertheless, we would like to use a similar technique, and construct a graph which will allow us to reason about independencies among the set of counterfactual variables in all submodels mentioned in $\gamma$.

The first attempt to construct such a graph was made in [Balke & Pearl, 1994a] where a *twin network graph* was constructed for $\gamma$ which mention exactly two submodels. The twin network graph consisted of two submodel graphs which shared exogenous variables $\mathbf{U}$.

One problem with the twin network graph, of course, is the restriction to two possible worlds. It can easily come to pass that a counterfactual query of interest would involve three or more worlds. For instance, we might be interested in how likely the patient would be to have a symptom $Y$ given a certain dose $x$ of drug $X$, assuming we know that the patient has taken dose $x'$ of drug $X$, dose $d$ of drug $D$, and we know how an intermediate symptom $Z$ responds to treatment $d$. This would correspond to the query $P(y_x|x', z_d, d)$, which mentions three worlds, the original model $M$, and the submodels $M_d, M_x$.

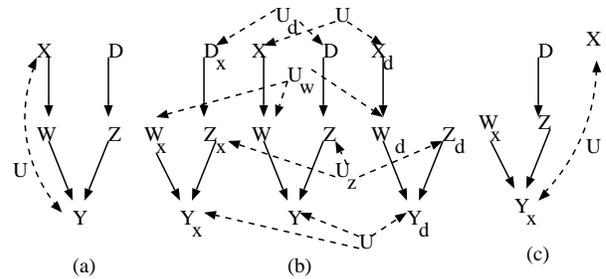

Figure 1: Nodes fixed by actions not shown. (a) Graph $G$. (b) Parallel worlds graph for $P(y_x|x', z_d, d)$ (the two nodes denoted by $U$ are the same). (c) Counterfactual graph for $P(y_x|x', z_d, d)$.

This problem is easy to tackle – we simply add more than two submodel graphs, and have them all share the same $\mathbf{U}$ nodes. This simple generalization of the twin network model was considered in [Avin, Shpitser, & Pearl, 2005], and was called there the parallel worlds graph. Fig. 1 shows the original causal graph and the parallel worlds graph for $\gamma = y_x \wedge x' \wedge z_d \wedge d$.

The other problematic feature of the twin network graph, which is inherited by the parallel worlds graph, is that multiple nodes can sometimes correspond to the same random variable. For example in Fig. 1 (b), the variables $Z$ and $Z_x$ are represented by distinct nodes, although it's easy to show that since $Z$ is not a descendant of $X$, $Z = Z_x$. These equality constraints among nodes can make the d-separation criterion misleading if not used carefully. For instance, $Y_x \not\perp Z|Z_x$ even though using d-separation in the parallel worlds graph suggests the opposite. To handle this problem, we use the following lemma which will tell us when variables from different submodels are in fact the same.

**Lemma 2** *Let $G$ be a causal diagram with $\mathbf{z}$ observed and $\mathbf{x}$ fixed. Then in all model inducing $G$ where nodes $\alpha, \beta$ share both the same functional mechanism and the same exogenous parents $U$, $\alpha, \beta$ are the same random variable if all their corresponding parents are either shared or attain the same value (either by intervention or observation).*

*Proof:* This follows from the fact that variables in a causal model are functionally determined from their parents. □

The parallel worlds graph can be thought of as a causal diagram for a special kind of causal model where some distinct nodes share the same functions. Using Lemma 2 as a guide, we want to modify such a diagram to rid ourselves of duplicate nodes, while at the same time ridding $\gamma$ of syntactically distinct variables which represent the same counterfactual variable. Since we need to establish sameness for parents before children, we apply Lemma 2 inductively starting with the root nodes.

If two nodes are established to be the same, we want to specify the rule for merging them in the graph. This rule work as we would expect. If two nodes (say corresponding to $Y_\mathbf{x}, Y_\mathbf{z}$) are established to be the same in $G$, they are merged into a single node which inherits all the children of the original two. These two nodes either share their parents (by induction) or their parents attain the same values. If a given parent is shared, it becomes the parent of the new node. Otherwise, we pick one of the parents arbitrarily to become the parent of the new node. The soundness of this operation is simple to establish.

**Lemma 3** *Let $M$ be a causal model with $\mathbf{z}$ observed, and $\mathbf{x}$ fixed such that Lemma 2 holds for $\alpha, \beta$. Let $M'$ be a causal model obtained from $M$ by merging $\alpha, \beta$ into a new node $\omega$, which inherits all parents and the functional mechanism of $\alpha$. All children of $\alpha, \beta$ in $M'$ become children of $\omega$. Then $M, M'$ agree on any distribution consistent with $\mathbf{z}$ being observed and $\mathbf{x}$ being fixed.*

*Proof:* This is a direct consequence of Lemma 2. □

The new node $\omega$ we obtain from Lemma 3 can be thought of as a new counterfactual variable. What should be its action (subscript)? Intuitively, it is those fixed variables which are ancestors of $\omega$ in the graph $G'$ of $M'$. Formally the subscript is $\mathbf{w}$, where $\mathbf{W} = An(\omega)_{G'} \cap \mathbf{sub}(\gamma)$, where the $\mathbf{sub}(\gamma)$ corresponds to those nodes in $G'$ which correspond to subscripts in $\gamma$. Since we replaced $\alpha, \beta$ by $\omega$, we replace any mention of $\alpha, \beta$ in our given counterfactual query $P(\gamma)$ by $\omega$. Note that since $\alpha, \beta$ are the *same*, their value assignments must be the same (say equal to $y$). The new counterfactual $\omega$ inherits this assignment.

We summarize the inductive applications of Lemma 2, and 3 by the **make-cg** algorithm, which takes $\gamma$ and $G$ as arguments, and constructs a version of the parallel worlds graph without duplicate nodes. We call the resulting structure the *counterfactual graph* of $\gamma$, and denote it by $G_\gamma$. The algorithm is shown in Fig. 2.

Note that there are three additional subtleties in **make-cg**. The first is that if variables $Y_\mathbf{x}, Y_\mathbf{z}$ were judged to be the same by Lemma 2, but $\gamma$ assigns them different values, this implies that $P(\gamma) = 0$. The second is that due to the factorization properties of causal graphs if we are interested in identifiability of $P(\gamma)$, we can restrict ourselves to the ancestors of $\gamma$ in $G'$ [Tian, 2002]. Finally, because the al-gorithm can make an arbitrary choice picking a parent of $\omega$ each time Lemma 3 is applied, both the counterfactual graph $G'$, and the corresponding modified counterfactual $\gamma'$ are not unique. This does not present a problem, however, as any such graph is acceptable for our purposes.

It's straightforward to verify that applying **make-cg** to the causal graph in Fig. 1 (a) and $\gamma = y_x \wedge z_d \wedge x' \wedge d$, one of the graphs that can be obtained is one in Fig. 1 (c).

## 4 Identification of Counterfactual Queries

Having constructed a graphical representation of worlds mentioned in counterfactual queries, we can turn to identification. We construct two algorithms for this task, the first is called **ID*** and works for unconditional queries, while the second, **IDC***, works on queries with counterfactual evidence and calls the first as a subroutine. These are shown in Fig. 2.

These algorithms make use of the following notation: $\mathbf{sub}(.)$ returns the set of subscripts, $\mathbf{var}(.)$ the set of variables, and $\mathbf{ev}(.)$ the set of values (either set or observed) appearing in a given counterfactual, while $\mathbf{val}(.)$ is the value assigned to a given counterfactual variable. $C(G')$ is the set of C-components, and $V(G')$ is the set of observable nodes of $G'$. Following [Pearl, 2000a], $G'_{\overline{y_\mathbf{x}}}$ is the graph obtained from $G'$ by removing all outgoing arcs from $Y_\mathbf{x}$; $\gamma'_{y_\mathbf{x}}$ is obtained from $\gamma'$ by replacing all descendant variables $W_\mathbf{z}$ of $Y_\mathbf{x}$ in $\gamma'$ by $W_{\mathbf{z},y}$. A counterfactual $\mathbf{s_r}$, where $\mathbf{s, r}$ are value assignments to sets of nodes, represents the event "the node set $\mathbf{S}$ attains values $\mathbf{s}$ under intervention $do(\mathbf{r})$."

We illustrate the operation of these algorithms by considering the identification of a query $P(y_x|x', z_d, d)$ considered in the previous section. Since $P(x', z_d, d)$ is not inconsistent, we proceed to construct the counterfactual graph on line 2. Suppose we produce the graph in Fig. 1 (c), where the corresponding modified query is $P(y_x|x', z, d)$. Since $P(y_x, x', z, d)$ is not inconsistent we proceed to the next line, which moves $z, d$ (with $d$ being redundant due to graph structure) to the subscript of $y_x$, to obtain $P(y_{x,z}|x')$. Finally, we call **ID*** with the query $P(y_{x,z}, x')$. The first interesting line is 6, where the query is expressed as $\sum_w P(y_{x,z,w}, x')P(w_x)$. Note that $x$ is redundant in the first term, so a recursive call reaches line 9 with $P(y_{z,w}, x')$, which is identifiable as $P_{z,w}(y, x')$ from $P_*$. The second term is trivially identifiable as $P_x(w)$, which means our query is identifiable as $P' = \sum_w P_{z,w}(y, x')P_x(w)$, and the conditional query is equal to $P'/P'(x')$.

When considering the soundness of our algorithms, the key observation is that the counterfactual graph which is output by **make-cg** is a causal diagram for a particular causal model. Thus, all the theorems that have been developed for ordinary causal models work for the counterfactual graph.

function **make-cg**$(G, \gamma)$
INPUT: $G$ a causal diagram, $\gamma$ a conjunction of counterfactual events
OUTPUT: A counterfactual graph $G_\gamma$, and either a set of events $\gamma'$ s.t. $P(\gamma') = P(\gamma)$ or **INCONSISTENT**

1. Construct a submodel graph $G_{\mathbf{x}_i}$ for each action $do(\mathbf{x}_i)$ mentioned in $\gamma$. Construct $G'$ by having all such graphs share their corresponding $U$ nodes.

2. Let $\pi$ be a topological ordering of nodes in $G'$. Apply Lemmas 2 and 3, in order $\pi$, to each node pair $\alpha, \beta$ sharing functions. If at any point $\mathbf{val}(\alpha) \neq \mathbf{val}(\beta)$, but $\alpha = \beta$ by Lemma 2, return $G'$, **INCONSISTENT**.

3. return $(An(\gamma')_{G'}, \gamma')$.

function **ID*** $(G, \gamma)$
INPUT: $G$ a causal diagram, $\gamma$ a conjunction of counterfactual events
OUTPUT: an expression for $P(\gamma)$ in terms of $P_*$ or **FAIL**

1. if $\gamma = \emptyset$, return 1

2. if $(\exists x_{x'..} \in \gamma)$, return 0

3. if $(\exists x_{x..} \in \gamma)$, return **ID***$(G, \gamma \setminus \{x_{x..}\})$

4. $(G', \gamma') = $ **make-cg**$(G, \gamma)$

5. if $\gamma' = $ **INCONSISTENT**, return 0

6. if $C(G') = \{S^1, ..., S^k\}$,
   return $\sum_{\mathbf{V}(G') \setminus \gamma} \prod_i $ **ID***$(G, s^i_{\mathbf{v}(G') \setminus s^i})$

7. if $C(G') = \{S\}$ then,

   8. if $(\exists \mathbf{x}, \mathbf{x}')$ s.t. $\mathbf{x} \neq \mathbf{x}', \mathbf{x} \in \mathbf{sub}(S), \mathbf{x}' \in \mathbf{ev}(S)$, throw **FAIL**
   9. else, let $\mathbf{x} = \bigcup \mathbf{sub}(S)$
      return $P_\mathbf{x}(\mathbf{var}(S))$

function **IDC*** $(G, \gamma, \delta)$
INPUT: $G$ a causal diagram, $\gamma, \delta$ conjunctions of counterfactual events
OUTPUT: an expression for $P(\gamma | \delta)$ in terms of $P_*$, **FAIL**, or **UNDEFINED**

1. if **ID***$(G, \delta) = 0$, return **UNDEFINED**

2. $(G', \gamma' \wedge \delta') = $ **make-cg**$(G, \gamma \wedge \delta)$

3. if $\gamma' \wedge \delta' = $ **INCONSISTENT**, return 0

4. if $(\exists y_\mathbf{x} \in \delta')$ s.t. $(Y_\mathbf{x} \perp\!\!\!\perp \gamma')_{G'_{\overline{y_\mathbf{x}}}}$,
   return **IDC***$(G, \gamma'_{y_\mathbf{x}}, \delta' \setminus \{y_\mathbf{x}\})$

5. else, let $P' = $ **ID***$(G, \gamma \wedge \delta)$. return $P'/P'(\delta)$

Figure 2: Counterfactual identification algorithms.

Thus, we reproduce a number of definitions and lemmas which hold for causal models which will help us in our proof.

**Definition 2 (c-component)** $G$ is a C-component if any two nodes $X, Y$ in $G$ are connected by a path where no observable node on the path has any outgoing arrows in the path. (such a path is called a confounding path).

C-components partition a causal diagram into a set of fragments where the distribution corresponding to each fragment is identifiable.

**Lemma 4** *For any $G$ and any effect $P_\mathbf{x}(\mathbf{y})$, $P_\mathbf{x}(\mathbf{y}) = \sum_{\mathbf{v} \setminus (\mathbf{y} \cup \mathbf{x})} \prod_i P_{\mathbf{v} \setminus s_i}(s_i)$, where $\{S_1, ..., S_k\}$ is the set of C-components of $G \setminus \mathbf{X}$.*

*Proof:* See [Tian, 2002], [Shpitser & Pearl, 2006b]. □

The truly new operation specific to identification in $P_*$ appears in line 9. We justify this operation with the following lemma.

**Lemma 5** *If the preconditions of line 7 are met, $P(S) = P_\mathbf{x}(\mathbf{var}(S))$, where $\mathbf{x} = \bigcup \mathbf{sub}(S)$.*

*Proof:* Let $\mathbf{x} = \bigcup \mathbf{sub}(S)$. Since the preconditions are met, $\mathbf{x}$ does not contain conflicting assignments to the same variable, which means $do(\mathbf{x})$ is a sound action in the original causal model. Note that for any variable $Y_\mathbf{w}$ in $S$, any variable in $(Pa(S) \setminus S) \cap An(Y_\mathbf{w})_S$ is already in $\mathbf{w}$, while any variable in $(Pa(S) \setminus S) \setminus An(Y_\mathbf{w})_S$ can be added to the subscript of $Y_\mathbf{w}$ without changing the variable. Since $Y \cap \mathbf{X} = \emptyset$ by assumption, $Y_\mathbf{w} = Y_\mathbf{x}$. Since $Y_\mathbf{w}$ was arbitrary, our result follows. □

**Theorem 1** *If **ID*** *succeeds, the expression it returns is equal to $P(\gamma)$ in a given causal graph.*

*Proof:* The first line merely states that the probability of an empty conjunction is 1, which is true by convention. Lines 2 and 3 follow by the Axiom of Effectiveness [Galles & Pearl, 1998]. The soundness of **make-cg** has already been established in the previous section, which implies the soundness of line 4. Line 6 follows by Lemma 4, and line 9 by Lemma 5. □

The soundness of **IDC*** is also fairly straightforward.

**Theorem 2** *If **IDC*** *does not output **FAIL**, the expression it returns is equal to $P(\gamma|\delta)$ in a given causal graph, if that expression is defined, and **UNDEFINED** otherwise.*

*Proof:* [Shpitser & Pearl, 2006a] shows how an operation similar to line 4 is sound by rule 2 of do-calculus [Pearl, 1995] when applied in a causal diagram. But we know that the counterfactual graph is just a causal diagram for a model where some nodes share functions, so the same reasoning applies. The rest is straightforward. □

## 5 Completeness

We would like to show completeness of **ID\*** and **IDC\***. To do so, we show non-identifiability in increasingly complex graph structures, until we finally encompass all situations where **ID\*** and **IDC\*** fail. Since we will be making heavy use of Lemma 1, we first prove a utility lemma that makes constructing counterexamples which agree on $P_*$ easier.

**Lemma 6** *Let $G$ be a causal graph partitioned into a set $\{S_1, ..., S_k\}$ of C-components. Then two models $M_1, M_2$ which induce $G$ agree on $P_*$ if and only if their submodels $M^1_{\mathbf{v}\setminus s_i}, M^2_{\mathbf{v}\setminus s_i}$ agree on $P_*$ for every C-component $S_i$, and value assignment $\mathbf{v} \setminus s_i$.*

*Proof:* This follows from C-component factorization: $P(\mathbf{v}) = \prod_i P_{\mathbf{v}\setminus s_i}(s_i)$. This implies that for every $do(\mathbf{x})$, $P_{\mathbf{x}}(\mathbf{v})$ can be expressed as a product of terms $P_{\mathbf{v}\setminus(s_i\setminus\mathbf{x})}(s_i \setminus \mathbf{x})$, which implies the result. □

The simplest non-identifiable counterfactual graph is the so called 'w-graph' [Avin, Shpitser, & Pearl, 2005], as the following lemma shows.

**Lemma 7** *Assume $X$ is a parent of $Y$ in $G$. Then $P_*, G \not\vdash_{id} P(y_x, y'_{x'}), P(y_x, y')$ for any value pair $y, y'$.*

*Proof:* See [Avin, Shpitser, & Pearl, 2005]. □

Intuitively, the problem with the 'w-graph' is that a variable $X$ is treated inconsistently in different worlds, while at the same time variables derived from $Y$ share the background context $U$, and $X$ is a direct parent of these variables. This means that it is not possible to use independence information to reconcile the inconsistency. This suggests the following generalization.

**Lemma 8** *Assume $G$ is such that $X$ is a parent of $Y$ and $Z$, and $Y$ and $Z$ are connected by a bidirected path with observable nodes $W^1, ..., W^k$ on the path. Then $P_*, G \not\vdash_{id} P(y_x, w^1, ..., w^k, z_{x'}), P(y_x, w^1, ..., w^k, z)$ for any value assignments $y, w^1, ..., w^k, z$.*

*Proof:* We construct two models with graph $G$ as follows. In both models, all variables are binary, and $P(\mathbf{U})$ is uniform. In $M^1$, each variable is set to the bit parity of its parents. In $M^2$, the same is true except $Y$ and $Z$ ignore the values of $X$. To prove that the two models agree on $P_*$, we use Lemma 6. Clearly the two models agree on $P(X)$. To show that the models also agree on $P_x(\mathbf{V} \setminus \mathbf{X})$ for all values of $x$, note that in $M_2$ each value assignment over $\mathbf{V} \setminus \mathbf{X}$ with even bit parity is equally likely, while no assignment with odd bit parity is possible. But the same is true in $M^1$ because any value of $x$ contributes to the bit parity of $\mathbf{V} \setminus \mathbf{X}$ exactly twice. The agreement of $M^1_x, M^2_x$ on $P_*$ follows by the graph structure of $G$.

To see that the result is true, we note firstly that $P(\Sigma_i W^i + Y_x + Z_{x'} \pmod 2) = 1) = P(\Sigma_i W^i + Y_x + Z \pmod 2) =$ 1) $= 0$ in $M^2$, while the same probabilities are positive in $M^1$, and secondly that in both models distributions $P(y_x, w^1, ..., w^k, z_{x'})$ and $P(y_x, w^1, .., w^k, z)$ are uniform. Note that the proof is easy to generalize for positive $P_*$ by adding a small probability for $Y$ to flip its normal value. □

To extend our results to more complex graph structures we need lemmas that allow us to make changes to the causal graph that preserve non-identification. It should be noted that versions of the following two lemmas also hold for identifying causal effects from $P$.

**Lemma 9 (contraction lemma)** *Assume $P_*, G \not\vdash_{id} P(\gamma)$. Let $G'$ be obtained from $G$ by merging some two nodes $X, Y$ into a new node $Z$ where $Z$ inherits all the parents and children of $X, Y$, subject to the following restrictions:*

- *The merge does not create cycles.*

- *If $(\exists w_s \in \gamma)$ where $x \in s$, $y \notin s$, and $X \in An(W)_G$, then $Y \notin An(W)_G$.*

- *If $(\exists y_s \in \gamma)$ where $x \in s$, then $An(X)_G = \emptyset$.*

- *If $(Y_\mathbf{w}, X_\mathbf{s} \in \gamma)$, then $\mathbf{w}$ and $\mathbf{s}$ agree on all variable settings.*

*Assume $|X| \times |Y| = |Z|$ and there's some isomorphism $f$ assigning value pairs $x, y$ to a value $f(x, y) = z$. Let $\gamma'$ be obtained from $\gamma$ as follows. For any $w_\mathbf{s} \in \gamma$:*

- *If $W \notin \{X, Y\}$, and values $x, y$ occur in $\mathbf{s}$, replace them by $f(x, y)$.*

- *If $W \notin \{X, Y\}$, and the value of one of $X, Y$ occur in $\mathbf{s}$, replace it by some $z$ consistent with the value of $X$ or $Y$.*

- *If $X, Y$ do not occur in $\gamma$, leave $\gamma$ as is.*

- *If $W = Y$ and $x \in \mathbf{s}$, replace $w_\mathbf{s}$ by $f(x, y)_{\mathbf{s}\setminus\{x\}}$.*

- *otherwise, replace every variable pair of the form $Y_\mathbf{r} = y, X_\mathbf{s} = x$ by $Z_{\mathbf{r},\mathbf{s}} = f(x, y)$.*

*Then $P_*, G' \not\vdash_{id} P(\gamma')$.*

*Proof:* Let $Z$ be the Cartesian product of $X, Y$, and fix $f$. We want to show that the proof of non-identification of $P(\gamma)$ in $G$ carries over to $P(\gamma')$ in $G'$.

We have four types of modifications to variables in $\gamma$. The first clearly results in the same counterfactual variable. For the second, due to the restrictions we imposed, $w_\mathbf{z} = w_{\mathbf{z},y,x}$, which means we can apply the first modification.

For the third, we have $P(\gamma) = P(\delta, y_{x,\mathbf{z}})$. By our restrictions, and rule 2 of do-calculus [Pearl, 1995], this is equal to $P(\delta, y_{\mathbf{z}}|x_{\mathbf{z}})$. Since this is not identifiable, then neither is $P(\delta, y_{\mathbf{z}}, x_{\mathbf{z}})$. Now it's clear that our modification is equivalent to the fourth.

The fourth modification is simply a merge of events consistent with a single causal world into a conjunctive event, which does not change the overall expression. □

Intuitively, the Contraction Lemma states that *knowing less* about the model, by having a coarser graph which considers two distinct nodes as one, will not help identification, as you would expect.

**Lemma 10 (downward extension lemma)** *Assume $P_*, G \not\vdash_{id} P(\gamma)$. Let $\{y^1_{\mathbf{x}^1}, ..., y^n_{\mathbf{x}^m}\}$ be a subset of counterfactual events in $\gamma$. Let $G'$ be a graph obtained from $G$ by adding a new child $W$ of $Y^1, ..., Y^n$. Let $\gamma' = (\gamma \setminus \{y^1_{\mathbf{x}^1}, ..., y^n_{\mathbf{x}^m}\}) \cup \{w_{\mathbf{x}^1}, ..., w_{\mathbf{x}^m}\}$, where $w$ is an arbitrary value of $W$. Then $P_*, G' \not\vdash_{id} P(\gamma')$.*

*Proof:* Let $M^1, M^2$ witness $P_*, G \not\vdash_{id} P(\gamma)$. We will extend these models to witness $P_*, G' \not\vdash_{id} P(\gamma')$. Since the function of a newly added $W$ will be shared, and $M^1, M^2$ agree on $P_*$ in $G$, the extensions will agree on $P_*$ by Lemma 6. We have two cases.

Assume there is a variable $Y^i$ such that $y^i_{\mathbf{x}^j}, y^i_{\mathbf{x}^k}$ are in $\gamma$. By Lemma 7, $P_*, G \not\vdash_{id} P(y^i_{\mathbf{x}^j}, y^i_{\mathbf{x}^k})$. Then let $W$ be a child of just $Y^i$, and assume $|W| = |Y^i| = c$. Let $W$ be set to the value of $Y^i$ with probability $1 - \epsilon$, and otherwise it is set to a uniformly chosen random value of $Y^i$ among the other $c - 1$ values. Since $\epsilon$ is arbitrarily small, and since $W_{\mathbf{x}^j}$ and $W_{\mathbf{x}^k}$ pay attention to the same $U$ variable, it is possible to set $\epsilon$ in such a way that if $P^1(Y^i_{\mathbf{x}^j}, Y^i_{\mathbf{x}^k}) \ne P^2(Y^i_{\mathbf{x}^j}, Y^i_{\mathbf{x}^k})$, however minutely, then $P^1(W_{\mathbf{x}^j}, W_{\mathbf{x}^k}) \ne P^2(W_{\mathbf{x}^j}, W_{\mathbf{x}^k})$.

Otherwise, let $|W| = \prod_i |Y^i|$, and let $P(W|Y^1, ..., Y^n)$ be an invertible stochastic matrix. Our result follows. □

Intuitively, the Downward Extension Lemma states that non-identification of causes translates into non-identification of effects (because the distribution over the latter can be in a one-to-one relationship with the distribution over the former). We are now ready to tackle the main results of the paper.

**Theorem 3** *ID\* is complete.*

*Proof:* We want to show that if line 8 fails, the original $P(\gamma)$ cannot be identified. There are two broad cases to consider. If $G_\gamma$ contains the w-graph, the result follows by Lemmas 7 and 10. If not, we argue as follows.

Fix some $X$ which witnesses the precondition on line 8. We can assume $X$ is a parent of some nodes in $S$. Assume no other node in $\mathbf{sub}(S)$ affects $S$ (effectively we delete all edges from parents of $S$ to $S$ except from $X$). Because the w-graph is not a part of $G_\gamma$, this has no ramifications on edges in $S$. Further, we assume $X$ has two values in $S$.

If $X \notin S$, fix $Y, W \in S \cap Ch(X)$. Assume $S$ has no directed edges at all. Then $P_*, G \not\vdash_{id} P(S)$ by Lemma 8. The result now follows by Lemma 10, and by construction of $G_\gamma$, which implies all nodes in $S$ have some descendant in $\gamma$.

If $S$ has directed edges, we want to show $P_*, G \not\vdash_{id} P(R(S))$, where $R(S)$ is the subset of $S$ with no children in $S$. We can recover this from the previous case as follows. Assume $S$ has no edges as before. For a node $Y \in S$, fix a set of childless nodes $\mathbf{X} \in S$ which are to be their parents. Add a virtual node $Y'$ which is a child of all nodes in $\mathbf{X}$. Then $P_*, G \not\vdash_{id} P((S \setminus \mathbf{X}) \cup Y')$ by Lemma 10. Then $P_*, G \not\vdash_{id} P(R(S'))$, where $S'$ is obtained from $S$ by adding edges from $\mathbf{X}$ to $Y$ by Lemma 9, which applies because no w-graph exists in $G_\gamma$. We can apply this step inductively to obtain the desired forest (all nodes have at most one child) $S$ while making sure $P_*, G \not\vdash_{id} P(R(S))$.

If $S$ is not a forest, we can simply disregard extra edges so effectively it is a forest. Since the w-graph is not in $G_\gamma$ this does not affect edges from $X$ to $S$.

If $X \in S$, fix $Y \in S \cap Ch(X)$. If $S$ has no directed edges at all, replace $X$ by a new virtual node $Y$, and make $X$ be the parent of $Y$. By Lemma 8, $P_*, G \not\vdash_{id} P((S \setminus x) \cup y_x)$. We now repeat the same steps as before, to obtain that $P_*, G \not\vdash_{id} P((R(S) \setminus x) \cup y_x)$ for general $S$. Now we use Lemma 9 to obtain $P_*, G \not\vdash_{id} P(R(S))$. Having shown $P_*, G \not\vdash_{id} P(R(S))$, we conclude our result by inductively applying Lemma 10. □

**Theorem 4** *IDC\* is complete.*

*Proof:* The difficult step is to show that after line 5 is reached, if $P_*, G \not\vdash_{id} P(\gamma, \delta)$ then $P_*, G \not\vdash_{id} P(\gamma|\delta)$. If $P_*, G \vdash_{id} P(\delta)$, this is obvious. Assume $P_*, G \not\vdash_{id} P(\delta)$. Fix the $S$ which witnesses that for $\delta' \subseteq \delta$, $P_*, G \not\vdash_{id} P(\delta')$. Fix some $Y$ such that a backdoor, i.e. starting with an incoming arrow, path exists from $\delta'$ to $Y$ in $G_{\gamma, \delta}$. We want to show that $P_*, G \not\vdash_{id} P(Y|\delta')$. Let $G' = An(\delta') \cap De(S)$.

Assume $Y$ is a parent of a node $D \in \delta'$, and $D \in G'$. Augment the counterexample models which induce counterfactual graph $G'$ with an additional binary node for $Y$, and let the value of $D$ be set as the old value plus $Y$ modulo $|D|$. Let $Y$ attain value 1 with vanishing probability $\epsilon$. That the new models agree on $P_*$ is easy to establish. To see that $P_*, G \not\vdash_{id} P(\delta')$ in the new model, note that $P(\delta')$ in the new model is equal to $P(\delta' \setminus D, D = d) * (1-\epsilon) + P(\delta' \setminus D, D = (d-1) \pmod{|D|}) * \epsilon$. Because $\epsilon$ is arbitrarily small, this implies our result. To show that $P_*, G \not\vdash_{id} P(Y = 1|\delta')$, we must show that the models disagree on $P(\delta'|Y = 1)/P(\delta')$. But to do this, we must simply find two consecutive values of $D$, $d, d+1 \pmod{|D|}$ such that $P(\delta' \setminus D, d+1 \pmod{|D|})/P(\delta' \setminus D, d)$ is different in the two models. But this follows from non-identification of $P(\delta')$.

If $Y$ is not a parent of $D \in G'$, then either it is further along on the backdoor path or it's a child of some node in $G'$. In case 1, we must construct the distributions along the backdoor path in such a way that if $P_*, G \not\vdash_{id} P(Y'|\delta')$ then $P_*, G \not\vdash_{id} P(Y|\delta')$, where $Y'$ is a node preceding $Y$ on the path. The proof follows closely the one in [Shpitser & Pearl, 2006a]. In case 2, we duplicate the nodes in $G'$ which lead from $Y$ to $\delta'$, and note that we can show non-identification in the resulting graph using reasoning in case 1. We obtain our result by applying Lemma 9. □

We conclude the paper by giving a graphical characterization of counterfactuals on which **ID\*** fails. Intuitively, the condition says that $P(\gamma)$ cannot be identified if actions and observations set variables in some C-component to conflicting values, and the conflicting variable is a parent of some node in the C-component. The properties of C-components then ensure that this conflict cannot be resolved using independence information in the model, resulting in non-identification.

**Theorem 5** *Let $G_\gamma, \gamma'$ be obtained from **make-cg**$(G, \gamma)$. Then $G, P_* \not\vdash_{id} P(\gamma)$ iff there exists a C-component $S \subseteq An(\gamma')_{G_\gamma}$ where some $X \in Pa(S)$ is set to $x$ while at the same time either $X$ is also a parent of another node in $S$ and is set to another value $x'$, or $S$ contains a variable derived from $X$ which is observed to be $x'$.*

*Proof:* This follows from Theorem 3 and the construction of **ID\***. □

## 6 Conclusions

In his critique of counterfactuals, [Dawid, 2000] argues that since counterfactuals cannot be directly tested, the use of counterfactual notation and counterfactual analysis should be avoided, lest it produces metaphysical or erroneous conclusions, unsubstantiated by the data. Our analysis proves the opposite [Pearl, 2000b]; only by taking counterfactual analysis seriously is one able to distinguish testable from untestable counterfactuals, then posit the more advanced question: what additional assumptions are needed to make the latter testable. We know, for example, that every counterfactual query is empirically identifiable in linear models. This implies that no counterfactual query is metaphysical if one can justify the assumption of linearity. Therefore, asking such queries in a non-linear context is not in itself metaphysical, but reduces to a mathematical question of whether the scientific knowledge at hand is sufficient for discerning the queries from the data.

In this paper we have provided a complete graphical criterion and associated algorithms for deciding whether an arbitrary counterfactual query of the form $P(Y_x|e)$ is discernible from experimental data when scientific knowledge is expressed in the form of an acyclic causal graph. Some counterfactual queries (e.g., the effect of binary treatment on the untreated patients) can be shown to be identifiable in general, with no additional assumptions needed. Others (e.g., the effect of a multi-valued treatment on the untreated), are identifiable only if the causal graph has a certain structure (e.g., Figure 1 (a)). The sensitivity of the results to graphical assumptions can be assessed using the bounding method of [Balke & Pearl, 1994a] which, again, is made feasible by the calculus of counterfactuals and their semantics.

Since all counterfactuals are empirically identifiable in linear systems, an interesting challenge would be to determine what properties of linear system can be given up without sacrificing empirical identifiability. Another interesting question is examine how these results can be carried over to the case of cyclic graphs.

### Acknowledgments

The authors thank anonymous reviewers for helpful comments. This work was supported in part by NSF grant #IIS-0535223, and NLM grant #T15 LM07356.